\documentclass[runningheads]{llncs}
\usepackage[T1]{fontenc}
\usepackage{graphicx}
\usepackage{graphicx}
\usepackage{xcolor}
\usepackage{amssymb}
\usepackage{amsmath}

\begin{document}
\title{Layout Stroke Imitation: A Layout Guided Handwriting Stroke Generation for Style Imitation with Diffusion Model}

\author{Sidra Hanif\inst{1}\and
Longin Jan Latecki\inst{2}}

\institute{Temple University, Philadelphia PA, USA\\
\email{\inst{1} sidra.haneef@yahoo.com, \inst{2}  latecki@temple.edu}} 
%
%\titlerunning{Abbreviated paper title}
% If the paper title is too long for the running head, you can set
% an abbreviated paper title here
%

% First names are abbreviated in the running head.
% If there are more than two authors, 'et al.' is used.
%

\maketitle              % typeset the header of the contribution
\begin{figure}[!ht]
    \centering
    \includegraphics[scale=0.4]{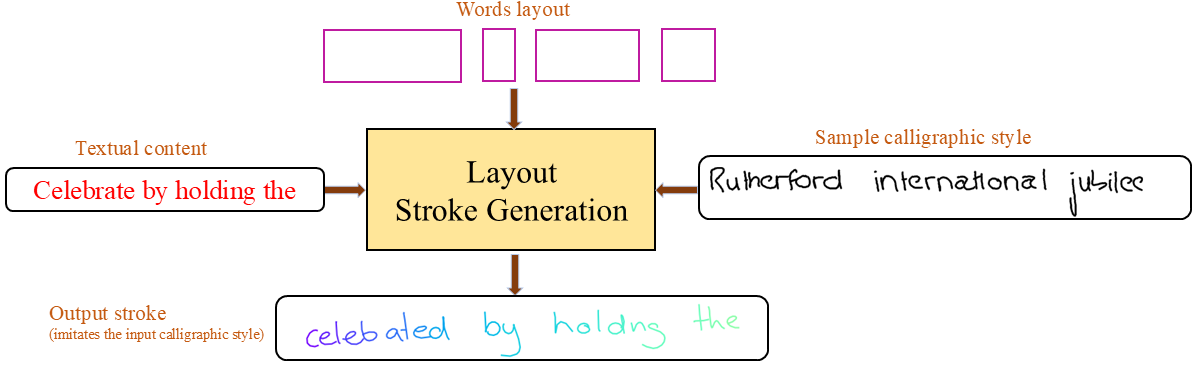}
     \caption{The input of sample calligraphic style, textual content, and words layout is fed into the proposed system for handwriting stroke generation. The output strokes imitate the calligraphic style for the given textual content and word layout.} 
     \label{fig:intro_img}
\end{figure}
\begin{abstract}
Handwriting stroke generation is crucial for improving the performance of tasks such as handwriting recognition and writer’s order recovery. In handwriting stroke generation, it is significantly important to imitate the sample calligraphic style. The previous studies have suggested utilizing the calligraphic features of the handwriting. However, they had not considered word spacing (word layout) as an explicit handwriting feature, which results in inconsistent word spacing for style imitation. Firstly, this work proposes multi-scale attention features for calligraphic style imitation. These multi-scale feature embeddings highlight the local and global style features. Secondly, we propose to include the words layout, which facilitates word spacing for handwriting stroke generation.
Moreover, we propose a conditional diffusion model to predict strokes in contrast to previous work, which directly generated style images. Stroke generation provides additional temporal coordinate information, which is lacking in image generation. Hence, our proposed conditional diffusion model for stroke generation is guided by calligraphic style and word layout for better handwriting imitation and stroke generation in a calligraphic style. 
Our experimentation shows that the proposed diffusion model outperforms the current state-of-the-art stroke generation and is competitive with recent image generation networks.

\end{abstract}
\section{Introduction} \label{sec:intro}

Handwriting stroke generation is an active research area that facilitates many subsequent tasks, such as handwriting recognition \cite{recog1} and writing order prediction \cite{pred}. 
In handwriting stroke generation, the desired calligraphic style is provided in the form of an image with a string of textual content. The system intends to learn by imitating the handwriting style for unseen textual content. Apart from mimicking the handwriting calligraphic style for stroke generation, the generated strokes need to create readable text. Despite the advancements of image generation models \cite{glide,dalle,cascade} for natural scene generation, precisely representing the calligraphic style of handwriting images in a generation network is still an open problem. In this work, we focus on handwriting stroke generation as opposed to the usually pursued handwriting image generation.
While it is a trivial task to generate an image from a given stroke sequence, the task of converting a handwriting image into a stroke sequence is very challenging. The input and desired output of our proposed system is shown in Fig \ref{fig:intro_img}.
%The handwriting stroke generation system is trained on a limited number of calligraphic style images and it aims to learn the style for any arbitrary text. 
\begin{table*}[]
\small
\hfill{}
\small
\begin{tabular}{ccccccc}

\hline
\multicolumn{1}{|c|}{\textbf{Method}}                                 & \multicolumn{1}{c|}{\textbf{Image}} & \multicolumn{1}{c|}{\textbf{Strokes}} & \multicolumn{1}{c|}{\textbf{Desired style}} & \multicolumn{1}{c|}{\textbf{words layout}} & \multicolumn{1}{c|}{\textbf{Long sentences}} \\ \hline
\multicolumn{6}{c}{\textbf{Handwriting stroke prediction}} \\ \hline
\multicolumn{1}{|c|}{Base LSTM \cite{bhunia2018}}    & \multicolumn{1}{c|}{}               & \multicolumn{1}{c|}{$\checkmark$}             & \multicolumn{1}{c|}{}                       & \multicolumn{1}{c|}{}                    &      \multicolumn{1}{c|}{}                                        \\ \hline
\multicolumn{1}{|c|}{Trace \cite{trace}}             & \multicolumn{1}{c|}{}               & \multicolumn{1}{c|}{$\checkmark$}             & \multicolumn{1}{c|}{}                       & \multicolumn{1}{c|}{}        & \multicolumn{1}{c|}{}                        \\ \hline
\multicolumn{1}{|c|}{U-STR \cite{u-trace}}           & \multicolumn{1}{c|}{}               & \multicolumn{1}{c|}{$\checkmark$}             & \multicolumn{1}{c|}{}                       & \multicolumn{1}{c|}{}        & \multicolumn{1}{c|}{$\checkmark$}            \\ \hline
\multicolumn{1}{|c|}{Inksight \cite{inksight}}           & \multicolumn{1}{c|}{}               & \multicolumn{1}{c|}{$\checkmark$}           & \multicolumn{1}{c|}{}                       & \multicolumn{1}{c|}{}        & \multicolumn{1}{c|}{$\checkmark$}            \\ \hline
\multicolumn{6}{c}{\textbf{Handwriting image generation}}                                                                                                                                                                                                                                                                               \\ \hline
\multicolumn{1}{|l|}{HiGAN+ \cite{higanplus}}        & \multicolumn{1}{c|}{$\checkmark$}   & \multicolumn{1}{l|}{}                                & \multicolumn{1}{c|}{$\checkmark$}           & \multicolumn{1}{c|}{}        & \multicolumn{1}{c|}{$\checkmark$}            \\ \hline
\multicolumn{1}{|c|}{One-DM \cite{oneshot}} & \multicolumn{1}{c|}{$\checkmark$}   & \multicolumn{1}{c|}{}                       & \multicolumn{1}{c|}{$\checkmark$}                       & \multicolumn{1}{c|}{}        & \multicolumn{1}{c|}{}                        \\ \hline
\multicolumn{1}{|c|}{VATr++ \cite{vatrpp}}               & \multicolumn{1}{c|}{$\checkmark$}   & \multicolumn{1}{c|}{}                        & \multicolumn{1}{c|}{$\checkmark$}                       & \multicolumn{1}{c|}{}        & \multicolumn{1}{c|}{}                        \\ \hline
\multicolumn{1}{|c|}{Wordstylist \cite{wordstylist}} & \multicolumn{1}{c|}{$\checkmark$}   & \multicolumn{1}{c|}{}                        & \multicolumn{1}{c|}{$\checkmark$}                       & \multicolumn{1}{c|}{}        & \multicolumn{1}{c|}{}                        \\ \hline
\multicolumn{6}{c}{\textbf{Handwriting stroke generation}}                                                                                                                                                                                                                                                                              \\ \hline
\multicolumn{1}{|c|}{Brush \cite{brush}}             & \multicolumn{1}{c|}{}               & \multicolumn{1}{c|}{$\checkmark$}                       & \multicolumn{1}{c|}{}                       & \multicolumn{1}{c|}{}                    & \multicolumn{1}{c|}{}                        \\ \hline
\multicolumn{1}{|c|}{Stroke diffusion \cite{vdiff}}  & \multicolumn{1}{c|}{}               & \multicolumn{1}{c|}{$\checkmark$}                      & \multicolumn{1}{c|}{}                       & \multicolumn{1}{c|}{}        & \multicolumn{1}{c|}{}                        \\ \hline
\multicolumn{1}{|c|}{Ours}                                            & \multicolumn{1}{c|}{}               & \multicolumn{1}{c|}{$\checkmark$}          & \multicolumn{1}{c|}{$\checkmark$}           & \multicolumn{1}{c|}{$\checkmark$}        & \multicolumn{1}{c|}{$\checkmark$}            \\ \hline
\end{tabular}
\caption{Capabilities of the previous and proposed methods for handwriting image and stroke generation.}
\label{tab:lit1}
\end{table*}
%Moreover, along with the above-mentioned conditions necessary for a feasible handwriting stroke generation system
The handwriting stroke prediction network is designed to learn the stroke trajectory from handwriting images. When given only handwriting image, the network predicts the sequence of strokes drawn by the writer to write the given text.  
%To estimate the stroke sequence of any arbitrary text in a desired style, we have to provide that text written in that particular calligraphic style. 
Since these networks \cite{trace,u-trace,brush,inksight} are not conditioned on arbitrary text, they cannot generate handwriting strokes for arbitrary text in a given calligraphic style. %unless provided with the text in the desired calligraphic style. 
In general, stroke prediction networks cannot generate text in unseen calligraphic styles because of their lack of textual conditioning and dependence on image features for both textual content and calligraphic style extraction. In our work, we present a handwriting stroke generation network that can generate strokes for an arbitrary text. The high-level workflow of the desired handwriting stroke prediction framework is shown in Fig. \ref{fig:intro_img}. The input is a calligraphic style in the form of an image and an arbitrary text in the form of a string, e.g., \textit{"celebrate by holding the"} and the word layout (bounding boxes of each word) of the arbitrary text. The system outputs a new handwriting stroke sequence that imitates the given calligraphic style.
%The current handwriting stroke prediction networks \cite{trace, u-trace,brush} cannot generate strokes for a given style without providing the handwriting image written in the desired calligraphic style.
On the other hand, recent advancements in generative models for natural images \cite{glide,dalle,cascade} have facilitated handwriting image generation as well. The natural images are conditioned on text prompts, whereas the handwriting images are conditioned on calligraphic style. However, the previous handwriting image generation methods use weak style learning networks to facilitate the readability of text \cite{higan,higanplus,vdiff,brush}. These networks are based on generative models and give less emphasis on calligraphic style features as compared to stronger textual features to increase the readability, which in turn reduces the generation model's capability to mimic calligraphic style. In contrast, our proposed multi-scale style feature extraction is specifically designed to provide distinctive calligraphic features for stroke generation for arbitrary textual content with additional word layout information to emphasize word spacing. 

%our work proposes to use multi-scale attention-based features for a handwritten image. 
We have noticed that by using strong multi-scale calligraphic style features, our method performs well on unseen style images (see Section \ref{sec:exp}) while keeping the readability intact.
%The existing approaches \cite{vdiff,brush} for handwriting stroke generation often are unable to provide distinctive features for different styles, which in turn reduces the generation model's capability to mimic calligraphic style.
%and grammatical correctness of the text .
Additionally, the image generation methods \cite{higan,higanplus,wordstylist} learn the background texture. However, for handwriting image generation, the handwriting style is more important than the background texture. 
One of the challenges of handwriting is word spacing. In the previous method, it is not considered as the explicit attribute of the handwriting. However, in our work, we explicitly provide word layout information along with textual content to emphasize word spacing in the generated strokes.      
%and grammatical correctness of the text. 
%\red{Mimic the handwriting style, Readability, Short/ long text, Reasonable time to train, Contain temporal stroke information for writing order
%Imitate handwriting calligraphic style for arbitrary text Able to generate strokes for unseen styles} With these models, generating images for arbitrary text and given calligraphic styles is possible. 
%One of the challenges that handwriting images offer is the spacing between words for different calligraphic styles. %Calligraphic style may include characters' shape and connectivity, font size, and writing tilt. 
%The variation in calligraphy styles makes handwriting image generation for a given textual string and calligraphic style difficult.
Moreover, the current methods generate one word at a time \cite{higan,higanplus,wordstylist,vatrpp,oneshot} and do not need to consider word spacing. However, our method can generate both words and lines of text with the desired word spacing. In Table \ref{tab:lit1}, we listed the capabilities of the previous methods with respect to the desired attributes of handwriting generation systems. 
%The handwriting stroke generation system is also preferred to work independently of the length of textual content. For instance, it is expected to generate readable strokes in a calligraphic style for short as well as longer text. 
%As we mentioned in Table \ref{tab:lit1}, our method can effectively generate strokes in any desired style for short and long sentences. Overall, handwriting stroke generation for arbitrary textual content and calligraphic styles is a challenging problem. Our work makes the following main contributions:
As shown in Table \ref{tab:lit1}, our method can effectively generate strokes in any desired style for short and long sentences. Overall, handwriting stroke generation for arbitrary textual content and calligraphic styles is a challenging problem.Our work makes the following main contributions:
\begin{itemize}
  \item We propose a conditional diffusion model for handwriting stroke generation. As opposed to handwriting generation as images, our model is small and efficient to train since we generate lower dimensional information (strokes) from the diffusion model.
  \item We propose multi-scale attention features to represent the calligraphic style. 
  %\item We also propose a character pair embedding to represent the connectivity between characters based on multi-scale features.
  \item We trained our diffusion model with word layout, which improves the word's spacing in the generated strokes.
  \item Our system can compete with image generation in terms of calligraphy style imitation via plotting strokes as images. 
  \item Our quantitative and qualitative results show our proposed method’s effectiveness and generalization ability.
\end{itemize}

\section{Related work}
We will briefly discuss the previous research methods for handwriting stroke prediction, handwriting image generation, and handwriting stroke generation.

\subsection{Handwriting strokes prediction}
%\red{We recommend that more online handwriting generation works[f,g,h,i] are disscussed in related work.}
%\red{[f] Deepwriting: Making digital ink editable via deep generative modeling, CHI, 2018. 
%[g] Generating handwriting via decoupled style descriptors, ECCV, 2020. 
%[h] Write Like You: Synthesizing Your Cursive Online Chinese Handwriting via Metric‐based Meta Learning, Computer Graphics Forum. 2021. 
%[i] Disentangling writer and character styles for handwriting generation, CVPR. 2023.} 
%are designed to predict the stroke trajectory from handwriting images \cite{trace, u-trace}. However, their design is limited to recovering the stroke trajectory from images, and they %For a long time, handwriting analysis, such as handwriting recognition \cite{recog1} and signature verification \cite{wor}, has been an active research area. The current STR architectures for English handwriting use lines of text \cite{trace,bhunia2018} or characters of alphabets \cite{multiseq,multiauto} as input.

Conventionally, handwriting stroke prediction methods \cite{wor,bioinspired} devised rule-based algorithms for word stroke prediction. 
%Stroke prediction has also benefited from the progress in deep neural networks. 
In recent years, \cite{multiseq,multiauto} introduced an attention mechanism to train the stroke prediction network for characters. These attention networks are trained on characters with L1-loss, which is challenging to train for words. 
%Similarly, %\cite{Japenese_char} employs an LSTM architecture with an attention layer and Gaussian mixture model trained with cross-entropy loss. However, it is limited to encoding only a single Japanese character. 
\cite{bhunia2018} introduced the first trainable LSTM architecture to learn strokes from Tamil scripts with Euclidean distance loss. It is difficult to apply to long words with multiple strokes, such as English handwriting. 
Recently, \cite{trace} presented a stroke trajectory recovery where LSTM is trained with a Dynamic Time Warping loss. However, this network does not go back to recover the stroke since it only predicts stroke in the forward direction. \cite{u-trace} improves on the discrepancy of \cite{trace} by including Chamfer distance loss and processing a shorter text to take advantage of DTW loss.  \cite{u-trace} can predict strokes for text in any orientation. However, it is also not able to predict strokes in the backward direction since LSTM models \cite{trace,u-trace} are restricted to only predicting strokes in the forward direction. 
\cite{inksight} alleviates this issue by using transformer-based architecture, but its capabilities are limited to stroke prediction from images, and it is not able to imitate calligraphic style.
Our system is able to imitate style and alleviate this issue by utilizing a generative diffusion model that does not limit stroke generation only in the forward direction.
%These methods are good enough for non-cursive calligraphic styles but they exhibit limitations for cursive calligraphic styles.
%All the methods mentioned above only predict strokes given the calligraphic style image.
\subsection{Handwriting image generation from style image}
The first few approaches for handwriting image generation conditioned on calligraphic style are based on GAN architecture. 
%Recently, several studies have proposed GAN architecture for handwriting image generation conditioned on calligraphic style. 
%\cite{alonso} proposed the first GAN-based architecture for synthesizing handwritten text images focusing on seen words only. 
\cite{ganwriting} proposed a few-shot architecture conditioned on the style for handwriting word generation. However, it is limited to synthesizing short words rather than long texts. Similarly, in AFFGANwriting \cite{affganwriting}, a style encoder based on VGG19 has been designed to extract multi-scale global and local features and fuse them for efficient calligraphic style. It results in generating much more realistic handwriting images.
%ScrabbleGAN \cite{scrabblegan} also synthesizes handwritten texts by concatenating all the letter tokens. It can be applied to any length of text but it does not generalize the calligraphic styles well and exhibits a lack of imitation ability.  
However, the GAN architectures are trained with multiple samples of images, such as 15 samples with the same calligraphic style. Recently, \cite{higan,higanplus} has been proposed to utilize a GAN-based framework for handwriting image generation. They can offer an advantage over the previous methods by using a single-word image as a sample, where \cite{higanplus} can produce realistic and readable output by taking advantage of patch discriminators, text recognition, and writer identification modules. %However, the generated images are slightly blurred. To improve the image quality and blur, \cite{higanplus} proposed to include a patch discriminator to the GAN architecture in \cite{higan}. 
%The local patches of the image are fed into a patch discriminator and trained to improve the blur in the generated images. 
However, it seems to generate images with a background even when there is no background texture in the style image (see Fig. \ref{fig:words_offline}). It is also not able to generalize unseen styles and requires a large memory to train because of several auxiliary networks. Since they generate words individually, there is no word spacing attribute being learned in these networks.
%They are not able to generate readable text because of their architectural limitations to validate the grammatical correctness of the generated text.
%utilizing the entire word embedding for style representation. However, this method cannot generate images for unseen words
Lately, \cite{HWT,vatr,vatrpp} utilize transformer encoder-decoder networks for handwriting image generation. %The transformer architecture \cite{HWT,vatr,vatrpp} shows an advantage over 
%However, it takes a long time to train but cannot imitate style despite including recognition and writer identification modules.
%\cite{vatr} proposed a transformer-based handwriting generation network, but they also require multiple style images to imitate the writer's handwriting style. 
% generates text with very low readability. 
%In general, transformer-based handwriting imitation networks take a long time to train and are difficult to converge. 
\cite{vatrpp} produces impressive handwriting imitation given the style template, but it generates either words or lines of text with no information about word spacing. It does not provide any stroke information of the generated text strokes. 
%with insufficient readability of generated text.%to learn style and background texture because of the auxiliary networks such as text recognition and writer identification. %Previous methods work are trained with multiple samples of images with the same calligraphic style.  

Recently, \cite{wordstylist,oneshot,diffpen} presented the latest diffusion model based image generation network without any auxiliary networks. The iterative learning of the diffusion model also requires a long time to train. Most of these networks are trained with writer class information; hence, it is hard to apply to an unseen style outside the dataset. Additionally, these methods   \cite{higan,higanplus,ganwriting,affganwriting,wordstylist,HWT,diffpen} only generate words and are not able to process long sentences because of the lack of word layout learning.
%Most of the previous methods \cite{higanplus,HWT,higan,wordstylist} include auxiliary networks such as text recognition and writer identification to improve the style features. However, these methods still exhibit limited applicability for unseen styles. 
Furthermore, for handwriting image generation, the learning network aims to generate images directly from calligraphic style features \cite{hgen1,hgen2,hgen3}, which requires a long time to train \cite{wordstylist}. On the other hand, handwriting stroke generation networks constitute fewer parameters to train and, therefore, can be computationally much faster to train \cite{vdiff,trace,u-trace}. 
In our work, we focus on stroke generation for style imitation with word layout fusion to mimic the global handwriting style.
%Moreover, handwriting image generation is restricted to generating only images without any temporal information, our methods contain richer temporal information for the writer's writing order. 
%Recent research attempts to propose a time-efficient solution for handwriting stroke generation.
% to embed the character shape, character pair connectivity, and overall writer
\subsection{Handwriting strokes generation from style image}
%\red{Unable to mimic handwringing style for arbitrary text, It is unable to generate unseen calligraphic styles, In LSTM architecture, predicted strokes cannot go back in time as it only predicts in forward direction }
The proposed approach belongs to handwriting stroke generation methods. We can generate strokes for an arbitrary textual string in any arbitrary calligraphic style. 
%The handwriting stroke generation has an advantage over the image generation. 
A stroke generation network has fewer parameters than an image generation network and requires less time to train. In previous approaches for stroke generation, \cite{brush} decouples the textual and style features from handwriting images but generates less illegible text because of the imperfect separation of textual and calligraphic features. Also, \cite{brush,disentangling,writelikeyou} synthesize text by decoupling style from template image; however, they struggle to generate readable text with reasonable character positioning and word spacing. 

Most recently, \cite{vdiff,hanif2024comprehensive} proposed a method to generate strokes from handwriting images in a given calligraphic style. It employs a diffusion model conditioned on text and style features to generate the stroke sequence for any arbitrary text in a given calligraphic style. It consists of a diffusion model without any auxiliary networks, which can be trained in a reasonable time. %However, it fails to generate strokes in the same calligraphic style and lacks diversity in the generated stroke styles. 
The drawback of this model is that they used mobilenet \cite{mobilenet} trained on natural images to extract the features from the handwriting images. Since the mobilenet does not represent handwriting features, it cannot mimic the calligraphic style of the image.  
% It results in irrelevant features for handwriting and does not represent the diversity in handwriting calligraphic styles 
For word spacing, \cite{deepwriting} predicts additional tokens such as end-of-character(\textit{eoc}) and beginning of word (\textit{bow}), so it can sometimes misrepresent the word spacing if the predicted value is incorrect. 

All the above-mentioned methods are trained for handwriting image generation without any input from word spacing or word layout.
Our work is inspired by the image generation from layout \cite{layout_text_to_image}, which is an emerging domain of natural image generation from the objects' layout in the image.
In the next section, we present our proposed diffusion model trained with multi-scale style features and guided with the word layout.

\section{Method}
%\red{The paper lacks a clear definition of its objectives at the beginning of the methods section. Specifically, it is unclear what the inputs and outputs are and how the strokes are described.}
In our work, we aim to generate strokes to mimic the writer's handwriting style from a single style image. The input for the system is an image of the handwriting style \textit{I$_{s}$}, the textual content \textit{T}, and the word layout \textit{L}. The system's output \textit{$G_S$} is the same textual content mimicked in the handwriting style of the input style image \textit{I$_{s}$}. 
The proposed method has three main components: multi-scale attention for style feature (Sec. \ref{ms}), text-layout encoder (Sec. \ref{mslay}), and a diffusion model (Sec. \ref{SDM}). 
A high-level block diagram of the proposed method is shown in Fig. \ref{fig:BD}.
%We describe the multi-scale attention (\textit{MS}) features for handwriting images in . In , we elaborate the design of the text-layout \textit{E}encoder. The overview of the diffusion model for handwriting stroke generation is described in .

%------------------------------------------------------------------------- 

\subsection{Multi-scale attention style features}\label{ms}
%\red{The proposed method appears to lack novelty, as it incorporates techniques previously utilized in similar contexts. For instance, AFFGANwriting[a] utilized multi-feature fusion for handwriting generation, while the employment of the Dynamic Time Warping algorithm (DTW) for training has already been demonstrated in [b] and [c]}
Generally, a multihead attention network processes images at a fixed resolution \cite{vit,svit}. However, handwriting images may constitute different font sizes, word spacing, and handwriting styles. To extract the style features at various granularity levels in the calligraphic style, we propose to compute the multihead attention features at multiple scales. Our mutihead attention network constitutes of three different positional embeddings, namely patch embedding, spatial embedding, and scale embedding. %Intuitively, multi-scale attention focuses on details in the original resolution images and on more global calligraphic style in the down-sampled variants.

\subsubsection{Patch embedding}
\begin{figure*}[!ht]
    \centering
    \includegraphics[scale=0.47]{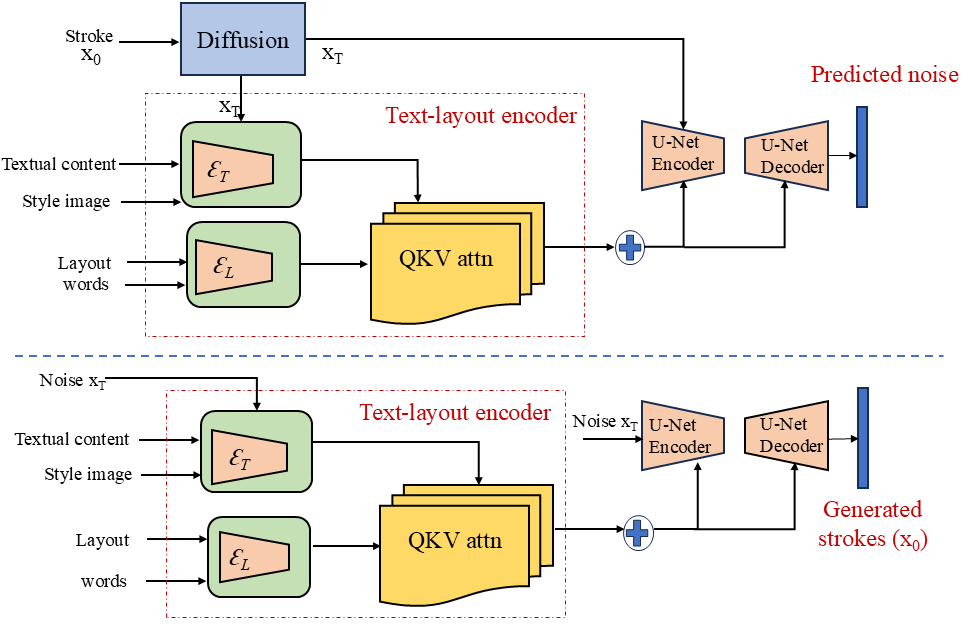}
     \caption{The overall block diagram of diffusion model with text-layout encoder module. Top module (Training) and bottom module (Inference).} 
     \label{fig:BD}
\end{figure*}
%Patches from different scale aggregate information across multiple scales and spatial locations 
%In handwriting images, feature representation from the character shape and global style plays a vital role in capturing the overall style for handwriting imitation. The proposed multi-scale handwriting style representation helps to capture the global and local style information. Patches from different scales enable the attention mechanism to aggregate information across multiple scales and spatial locations. 

The input for our multi-scale attention comprises the full-size image with height H (128), width W (1024), channel C (3), and two resized variants using a Gaussian kernel of size 96x768 and 64x512. The downsampled variants have height \textit{$h_{k}$}, width \textit{$w_{k}$}, channel \textit{C}, where \textit{k} = [1, 2] since we are using two resized variants. % The input image is a 128x1024 dimensional grayscale image. The two variants are down-sampled at 96x768 and 64x512 resolution. 
%Multi-scale attention tends to focus on character shape details in the original image resolution, however, the multi-scale attention at the two down-sampled variants focuses on a more global calligraphic style. 
The feature representation from downsampled images improves the quality of the feature's representation and makes them independent of the quality of the input handwriting image. 

 %We pad the image with zeros if the width or height is not multiples of \textit{P}. 
 
%Since all the patches from different scales are fed into the same patch embedding module, resulting in all the patches have different embedding.
%\blue{Since the patch embedding module focuses on computing the embedding of each patch, it assigns unique patch embedding to individual patches in each scale. Therefore, the visually similar patches  have different embeddings in each scale even though they are visually similar. However, the desired property for positional embedding is that the spatially close patches must have the same positional embedding irrespective that they belong to different scales}. 

We extract square patches of size $P \times P$ from each image in the multi-scale representation. The patch embedding module is intended to compute the embedding of each patch, assigning a unique embedding to every patch across different scales. Consequently, patches that look visually similar and are located in the same position may have different embeddings in each scale, despite their visual similarity. Yet, the ideal characteristic for positional embedding is that spatially proximate patches should share the same positional embedding, regardless of whether they belong to different scales. To satisfy this property, we describe a spatial embedding in the next section, which ensures that the spatially close patches in different scales have the same spatial embedding.

\begin{figure}[htbp]
  \begin{minipage}{0.48\textwidth}
   \centering
    \includegraphics[scale=0.3]{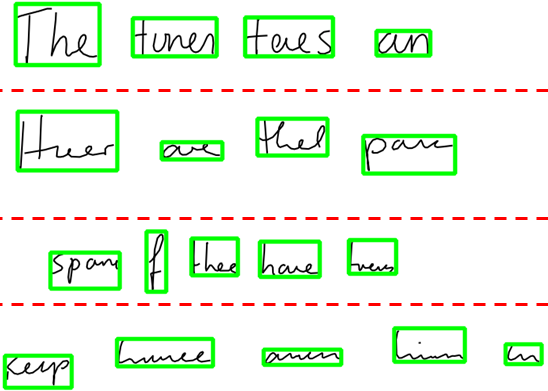}
     \caption{The word layout provided from word detection.} 
     \label{fig:wd}
  \end{minipage}%
  \hfill
  \begin{minipage}{0.48\textwidth}
    \centering
    \begin{tabular}{cc}
    \hline
    \multicolumn{2}{|c|}{Online writer ID}                                \\ \hline
    \multicolumn{1}{|c|}{Methhod}                                & \multicolumn{1}{c|}{\%acc} \\ \hline
    \multicolumn{1}{|c|}{Baseline \cite{vdiff}} & \multicolumn{1}{c|}{4.90}  \\ \hline
    \multicolumn{1}{|c|}{Multi-scale {[}4, 32{]}}                & \multicolumn{1}{c|}{\textbf{32.35}} \\ \hline
    
    \multicolumn{2}{|c|}{Offline writer ID}                                \\ \hline
    \multicolumn{1}{|c|}{Method}                                  & \multicolumn{1}{c|}{\%acc} \\ \hline
    \multicolumn{1}{|c|}{Baseline \cite{resnet}} & \multicolumn{1}{c|}{87.07} \\ \hline
    \multicolumn{1}{|c|}{Multi-scale {[}4,32{]}}                  & \multicolumn{1}{c|}{\textbf{95.60}} \\ \hline
    \end{tabular}
    \caption{The accuracy of writer ID for classification.}
    \label{fig:writerID}
  \end{minipage}
\end{figure}

%\red{.The figures in the paper are of low quality and are sometimes confusing. For instance, it is unclear what the arrow of patch embedding in Fig.3 represents. Additionally, it is ambiguous whether the embedding ( E ) in Fig.4 represents the feature of a single character or an entire word.}
%The sequence of patch embedding output from the patch encoder modules is concatenated to form a multiscale embedding sequence for the input image. 

\subsubsection{Spatial embedding}
%As we have mentioned before, we leverage patches from different scales for better style features for handwritten images. It also imposes an additional constraint on the positional embedding. 

As we mentioned before, patches from different scales corresponding to the same image portions should have the same spatial embedding. 
%The spatial embedding is required to follow the set of requirements such as 1) effectively encode the 2D spatial position of each patch to 1D sequence; 2) spatially close patches at different scales should have close spatial embedding; 3) efficient for computing the multiscale attention. On the other hand, traditional positional embedding assigns different patch embedding to each patch and is not able to align the spatially close patches from different scales.
%. However, the spatially close patches in different scales should have the same positional embedding.
Based on that, we utilize hash-based 2D spatial embedding (HSE). The patch at the location (row i, column j) is hashed to the corresponding element in a $G_{h} \times G_{w}$ grid, where each element in the grid is a D-dimensional embedding. \textit{HSE} is defined by a learnable matrix $T \in R^{G_{h} \times G_{w}\times D}$.
The input with resolution $H \times W$ is partitioned into $\frac{H}{P} \times \frac{W}{P}$ patches. For the patch at position (i, j), its spatial embedding is defined by the element at position ($t_{i}$, $t_{j}$) in T where
\begin{equation}
t_i=\frac{i \times G_h}{H / P}, t_j=\frac{j \times G_w}{W / P}
\end{equation}

%Each element in \textit{T} is a D-dimensional learnable embedding. 
The patch located at row \textit{i} column \textit{j} of the image is hashed to the corresponding element ($t_{i}$, $t_{j}$) of matrix \textit{T}. The Fig. \ref{fig:musiq} shows that the patch in original, scale 1, and scale 2 are pointing to the same location  ($t_{i}$, $t_{j}$) in grid G $\times$ G.
The D-dimensional spatial embedding $T_{{t}_{i}},_{{t}_{j}}$ is added to the patch embedding element-wisely as shown in Fig \ref{fig:musiq}.
To ensure alignment of patches across different scales, patches
located closely in the image but from different scales are
mapped to spatially close embeddings $T_{{t}_{i}},_{{t}_{j}}$, since \textit{i} and \textit{H}, as well as \textit{j} and \textit{W}, change proportionally to the resizing factors. The hash spatial embedding is inspired by \cite{musiq}.
%Patch locations from all scales are mapped to the same grid \textit{T} to align patches across scales. As a result, 
We selected the appropriate grid size through experimentation. As we know, handwriting sentence has a longer length than their height, so using the same grid size for width and height dimensions is not an appropriate choice. Therefore, we propose using the smaller grid size for height compared to its width. For the IAM-online \cite{IAM-online} dataset, we used [$(G_{h}, G_{w})$ ]= [4x32] grid size as an appropriate choice, where $G_{h}$ is a grid size for height and $G_{w}$ is a grid size of the width. 
% $G_{w}$ may result in a lot of collision between patches, making the model unable to distinguish spatially close patches. Larger $G_{w}$ requires large memory and may need more diverse resolutions to train. 
The smaller size of $G_{h}$ might result in overlapping patches; however, the larger values of $G_{h}$ would not be able to capture the local feature across the height dimension of the handwriting. %The block diagram of multi-scale attention with patch and spatial embedding is shown in Fig. \ref{fig:musiq}.
The implementation details of spatial embedding are given in Supplementary material Sec. 1.

\subsubsection{Scale embedding}

\begin{figure}[htbp]
  \begin{minipage}{0.48\textwidth}
    \includegraphics[scale=0.33]{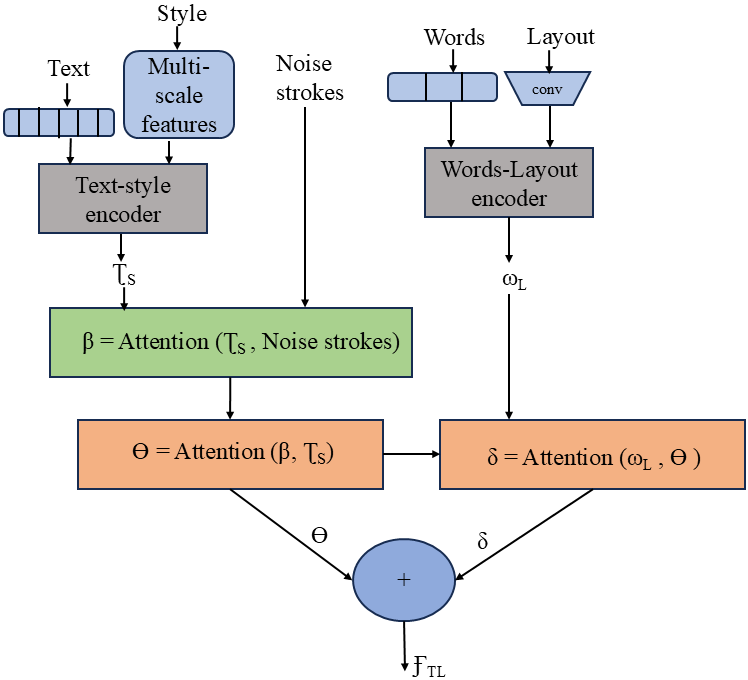}
     \caption{Block diagram of a text-layout encoder for stroke diffusion model.} 
     \label{fig:vdiff}
  \end{minipage}%
  \hfill
  \begin{minipage}{0.48\textwidth}
    \centering
   \includegraphics[scale=0.2]{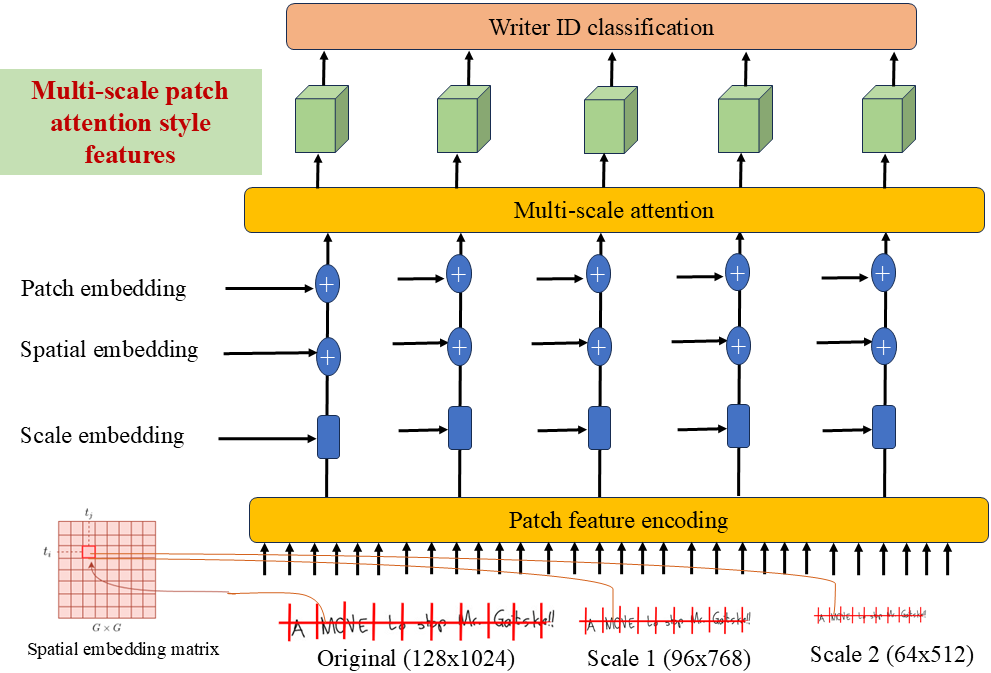}
     \caption{The architecture for multi-scale learning with patch, spatial, and scale embeddings with writer ID classification.} 
     \label{fig:musiq}
  \end{minipage}
\end{figure}

The spatial embedding satisfies the condition of assigning the same embedding to spatially close patches in different scales. However, it does not distinguish information coming from different scales. So, we define another embedding called scale embedding to help the attention model effectively distinguish information coming from different scales. 
%It aims to utilize style information across scales better since the spatial embedding does not make a distinction between patches from different scales.  
We define scale embedding as a learnable embedding Q $\in R^{(K+1)×D}$ for the input image and two downsampled variants. Following the spatial embedding, the first element $Q_{0} \in R^{D}$ is added element-wise to all the D-dimensional patch embeddings from the original image resolution. $Q_{k} \in R^{D}$ k = 1, 2 are also added element-wisely to all the patch embeddings from the downsampled variants as shown in Fig \ref{fig:musiq}.
The sum of patch embedding, spatial embedding, and scale embedding is fed into a multi-head attention network. We train our multi-scale attention network to identify writers from handwriting images. 
Figure \ref{fig:writerID} shows the accuracy of writer identification for online and offline handwriting images. 
We compare the proposed methods with baseline mobile net \cite{vdiff} for online handwriting and residual network \cite{resnet} for offline handwriting. We can see that the proposed multi-scale features outperform the baseline for both online and offline by a large margin. %Baseline methods are CNN networks Which have the disadvantage of extracting features at fixed image size. However, we extracted multi-scale features which improve the overall writer identification accuracy.
%The local patch features of dimensions 77x384 are extracted from the multi-scale attention network before the classification layer.%The 77 local patches each of dimensions 384 embeds the local and global style information of handwriting style. 
The local patches (77x384) with rich style information are used to train the diffusion model in Sec \ref{SDM}. To the best of our knowledge, the representation capabilities of multi-scale patch embedding, spatial embedding, and scale embedding have not been explored before to represent handwritten images' local character and global style features. Our experimentation in Sec. \ref{results} validates the effectiveness of these features for handwriting stroke generation. For the case of offline images, proposed multiscale features give the writer a classification accuracy of 95.60\%, and online images give 32.35\%. The accuracy of online images is reasonable because online images are significantly less diverse than the offline images. The previous work \cite{affganwriting} also utilized multi-scale feature fusion for style features to generate handwriting images. But, in contrast to our work, their features are fused from different levels of a convolutional network for the same image resolution. 
%In our work, we consider images at three scales. One at the original image resolution and two down-sampled variants with their aspect ratio preserved (ARP). 

%\red{The proposed style distance metrics are not entirely convincing. As shown in Table 2, the learned feature extractor achieves only 32\% accuracy for writer ID classification on online data. Given this low accuracy, it is questionable whether the features extracted by the model contain valid information for evaluation purposes.}

%-------------------------------------------------------------------------

\subsection{Text-layout encoder}\label{mslay}
%\red{Additionally, the definition of the style condition ( S ) in L565 is vague. For a given character, it is unclear what its corresponding character pair is in Sec. 3.2. This information is crucial for understanding the overall framework.}
The multi-scale (\textit{MS}) features introduced in the last section are effective in encoding a local character shape as well as the global handwriting style. For style \textit{S} conditioning, we first extract \textit{N} local patch features of each \textit{k} dimension from the handwriting image. The text embedding module embeds each character from the text string \textit{T} into \textit{K}-dimensional embedding $E$. The attention mechanism computes the attention of each text character embedding to each patch of the local style features. The embedding 
%Fig. \ref{fig:vdiff} shows the attention between style features and text sequence. The attention output is then added to the text sequence before passing through a feedforward network, which learns the compact text-style encoding. However, it lacks the word spacing information. 

%Fig. \ref{fig:vdiff}shows the compact text-style encoding based on local style patches and text embedding. It encodes each character into the writer style. For instance, each character in the text (\textit{sentence}) is utilized separately to attend to style features. 

In our work, we propose embedding word layout information into text-style features. By incorporating the words' positioning into the sentences, we aim to enhance the writer's overall style.
%In our work, we propose a comprehension style-text attention and strengthen the writer's style with the calligraphic style between character pairs as well. We refer to these style features as multi-scale character attention pairs (\textit{MSCAP}).

%In the more comprehension style-text attention, . For this purpose, we introduce a .
%Here, we get two sets of style features: one is character attention attn(\textit{S}, \textit{T}), and the other is character pair \textit{P}  attention attn(\textit{P}, \textit{T}).
The text attention with style features, which we called text-style \textit{$T_S$} embedding, is computed by the text-style encoder. It embeds the local character shape into a writer's style. 

However, the word layout attention with text-style improves the word spacing between words in generated Handwriting strokes. Finally, we sum up the text-style features and layout attention on text-style features and name it as text-layout embedding. In Fig \ref{fig:vdiff}, $\Gamma_{s}$ is a text-style and $\omega_{L}$ is a word-layout feature. $\beta$ computes the attention between $\Gamma_{s}$ and stroke sequence and provides it another attention layer, which computes attention of $\beta$ with $\Gamma_{s}$. Eventually, we compute the attention of $\Theta$ with $\omega_{L}$. Finally, we add $\Theta$ with $\delta$, which serves as a text-layout feature. 

We condition the diffusion model on the proposed text-layout features for handwriting stroke generation. The style is extracted from the handwritten image template, and the text content is the given arbitrary text that we intend to generate in the same style as the style template along with the word layout.
This method of attention between style features and text sequence is effective in conditioning the diffusion model for stroke prediction. 
Our feature extraction method surpasses the baseline \cite{mobilenet} for handwriting style features proposed in \cite{vdiff}. 
%The multi-scale features presented in Section \ref{ms} show promising results for encoding style features for diffusion model. The character shapes are effectively encoded in multi-scale features. 
%We also propose to compute the combination of two 
In Section \ref{sec:exp}, we demonstrate the effectiveness of our proposed text-layout features for handwriting imitation. To the best of our knowledge, this is the first study to explore the multi-scale handwriting style features and word layout with a diffusion model for handwriting stroke generation.

%\red{ The architecture of the diffusion model presented in Fig.5 appears to be incorrect. The inclusion of convolution and pooling operations in the diffusion process, as shown in the left-half of the network structure, is questionable. The diffusion process should simply add noise to the ground truth, which can be achieved through an analytic process as described in Eq.(3). }
\subsection{Stroke diffusion Model}\label{SDM}
Our diffusion model is conditioned on text-layout embedding from Sec. \ref{mslay}. We iteratively add the Gaussian noise to the ground truth stroke sequence. In general, the diffusion model employs Markov chains to add noise and disrupt the structure of input data, this step is called the diffusion process. In the reverse process, the model then learns to reverse the diffusion process and tries to reconstruct the original data, this process is called the denoising process \cite{sohl}. We presented the mathematical details of the diffusion model in the supplementary material in Sec 2.
%Note that our diffusion model consists of consecutive convolution layers and attention blocks as shown in Fig. \ref{fig:diff_model}.
%The attention block as shown in Fig. \ref{fig:diff_model} computes the attention between n stroke sequence and text-layout features. 
For training, the attention blocks as shown in Fig. \ref{fig:BD} condition the diffusion model on text-layout features and noisy ground truth stroke sequence. 
%The decoder part also includes a convolutional layer with upsampling layers which is designed to predict the noise scores. We consider a diffusion model \cite{diff1} as a score-based generative model, where instead of learning to model the energy function itself from latent distribution, we learn the score of the energy-based model as a neural network.
The success of the conditional diffusion model for image generation makes it a suitable technique for generating handwriting images conditioned on the writer's style for handwriting imitation. The image generation in the diffusion model is learned by iteratively adding small amounts of noise to an image and changing it into a random image during training. The model learns to reverse this process, generating realistic images by removing noise. The emerging image generation models are based on diffusion models \cite{diff1,diff2,diff3}. However, image generation is a computationally expensive process. Therefore, our work proposes to generate the stroke sequences using a diffusion model conditioned on the writer's style and textual content. It could be trained in a reasonable amount of time (see Table \ref{tab:arch}). It can also predict additional temporal information in the form of stroke sequence for the writer's handwriting which is not available for image generation \cite{higan,higanplus,wordstylist,HWT,vatr}.

\subsubsection{Inference}
During sampling, diffusion models iteratively remove the noise added in the diffusion process, by sampling $y_{t-1}$ for t = T, ... , 1. The stroke sequence $y_{t-1}$ at time step \textit{T-1} is computed with the equation below.
\begin{equation}
\mathrm{y}_{t-1}=\frac{1}{\sqrt{a_t}}\left(\mathrm{y}_t-\frac{\beta_t}{\sqrt{1-\bar{\alpha}_t}} \epsilon_\theta\left(\mathrm{y}_t, t\right)\right)+\sigma_t z
\end{equation}

where $z \sim \mathcal{N}(0, I)$ and $\alpha_{t}$ is a constant related to $\beta_{t}$. For our experiments, we used $\alpha_{t}^{2} = \beta_{t}$.

In our diffusion model, we sample uniform noise to provide an input stroke sequence to predict the stroke sequence in the desired style for given textual content and word layout. The diffusion model is provided with the text-layout embedding of the desired style from the image and textual content from a text string and bounding boxes of words from the words layout. In this way, we do not need the ground truth strokes in the inference phase, and the learned diffusion model effectively generates strokes from noise given text-layout embedding. During inference, we perform diffusion and denoising processes with the addition of a sampling process, as shown in Fig. \ref{fig:BD}. Our diffusion model design facilitates us to generate strokes for any arbitrary calligraphic style given any arbitrary text content.

\subsection{Loss function}\label{loss}

The output of our handwriting stroke generation $x$ is composed of a sequence of 
%\textit{N} vectors $x_{1} . . . x_{N}$. 
$N$ vectors $x_{1} \dots x_{N}$. Each vector in the sequence $x_i$ is composed of a real-valued pair, which represents the pen offset from the previous stroke in the x and
y direction along with a binary entry that has a value of 0 if the pen was writing the stroke and 1 otherwise. 
%_{n} \in R^{2}$ x ${0,1}
Each handwritten sequence is associated with a discrete character sequence \textit{C} describing the text. Each sequence is also associated with an offline image containing the writer’s style information, denoted by \textit{S}.
%There is one technical issue that needs to be addressed. The reverse process in Equation \ref{eq:3} is parameterized by a Gaussian distribution.
Since We cannot parameterize the binary variable representing whether the stroke was drawn by a Gaussian distribution as we did for the real-valued pen strokes. Instead, we parameterize it with a Bernoulli distribution.
For this purpose, we split each data point $x_{i}$ into two sequences $y_{i}$ and $d_{i}$ of equal length, with $y_{i}$ representing the real valued pen strokes, and $d_{i}$ representing whether the stroke was drawn. At each step t,
our model $d_{\theta}$ returns an estimate $\hat{d_i}$ of whether the pen was down. %\blue{$d_{\theta}$ and $\epsilon_{\theta}$ both are predicted by the same parameters of the diffusion model.} 
\begin{equation}
L_{\text{stroke}}(\theta) = \left\| \epsilon - \epsilon_\theta\left(y_t, c, s, \sqrt{\bar{\alpha}} \right) \right\|_2^2
\end{equation}
\begin{equation}
L_{\text {drawn }}(\theta)=-\mathrm{d}_0 \log \left(\hat{\mathrm{d}}_0\right)-\left(1-\mathrm{d}_0\right) \log \left(1-\hat{\mathrm{d}}_0\right)
\end{equation}
%Diffusion models are generally trained with diffusion loss, as mentioned above. However, recent attempts on image generation \cite{auxloss,contdiff,patchdiff} have shown that the additional loss would help the diffusion network to generate more realistic results.
%As in \cite{contdiff,patchdiff} a diffusion loss is applied on image patches to improve the quality of the generated image.
\section{Experiments}\label{sec:exp}
For our experimentation, we used lines of text from the IAM-online dataset \cite{IAM-online}. To evaluate our methods and compare them with previous works, we used the Inception Score (IS), Fréchet inception distance (FID), along with Peak signal-to-noise ratio (PNSR) and Mean Structure Similarity Index Method (MSSIM) matrices. The details of these metrics can be found in supplementary material Sec. 3

%We also compute the loss for predicting the start-of-stroke token $\epsilon_{sos}$. $\epsilon_{sos}$ should have the lowest value if the start-of-stroke token is predicted correctly.

%-------------------------------------------------------------------------

\begin{figure*}[!ht]
    \centering
    \includegraphics[scale=0.53]{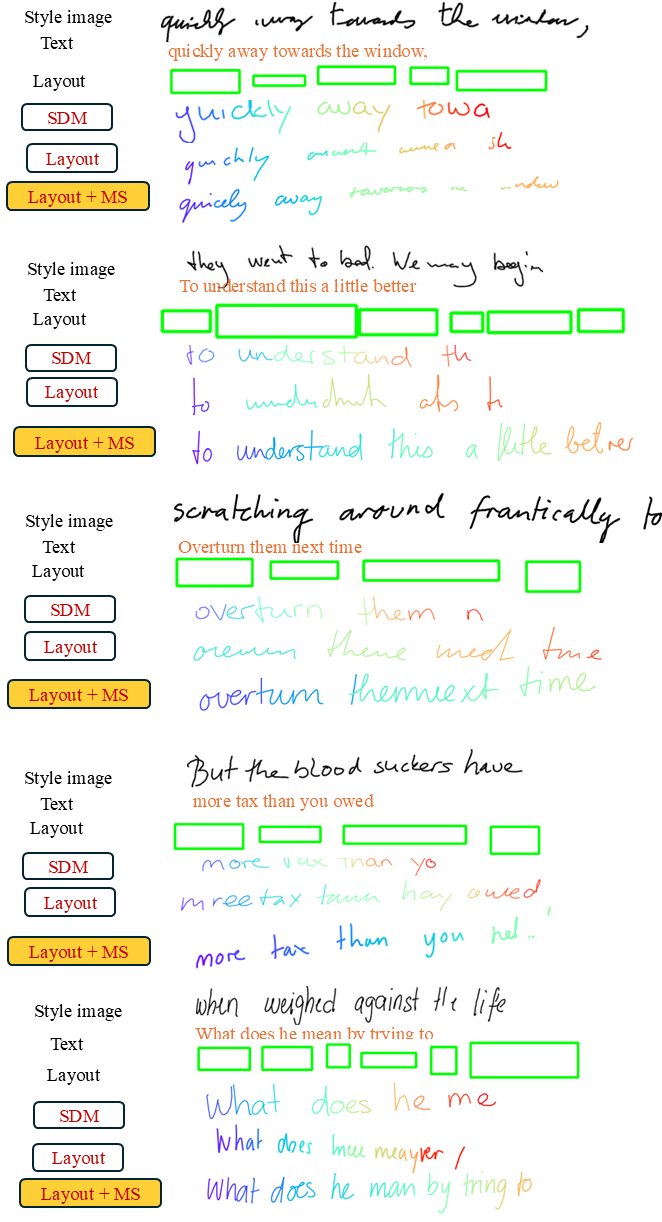}
     \caption{Visual comparison of the proposed method for stroke generation given the style image, text content, and word layout.} 
     \label{fig:bdiagram}
\end{figure*}
\begin{table*}[] 
\centering
\begin{tabular}{c|c|c|c|c}
\hline
Method            & IS $\uparrow$   & FID $\downarrow$ & PSNR  $\uparrow$ & MSSIM $\uparrow$ \\ \hline
HiGAN+ \cite{higanplus}         & 1.594       & 2.310         & 11.749 & 0.685    \\ \hline
wordstylist  \cite{wordstylist}     & 1.861          & 0.910          & 10.272         & 0.501                                \\ \hline
Stroke diffusion \cite{vdiff}       & 1.420 & 2.130   & 12.440 & 0.741   \\ \hline  
VATR++ \cite{vatrpp}       &1.513  & 1.397 &10.995 &  0.588   \\ \hline
One-DM \cite{oneshot}       & 1.656 & -  & 10.270& 0.244   \\ \hline
Layout + MS (Ours)   & 1.561 &  1.383 & 12.459 & 0.746    \\ \hline
\end{tabular}
\caption{Quantitative comparison of our method with state-of-the-art methods for handwriting imitation. The style input
is an online handwritten image from the IAM-offline dataset.}
\label{tab:overall}
\end{table*}
\subsection{Results}\label{results}
We utilized IAM-online \cite{IAM-online} dataset to train our diffusion network. It includes the images of handwritten text, the textual content in the image as a string of characters, and the x and y coordinates of the strokes with a pen-up and down information. We trained the word detection network \cite{hanif_detection} for the layout of words as described in Sec. \ref{mslay} and shown in Fig \ref{fig:wd}. We preferred the single stage network \cite{hanif_detection} than the two-stage network \cite{isearch} .
To evaluate the quality of image generation, we divide our analysis into two scenarios: online and offline input sample images. In the first scenario of online handwriting images, we provide the style image generated via stroke generation. These images have no texture and only contain black handwriting on a white background. Sample input style images are shown at the top of each example in Fig. \ref{fig:bdiagram}. 
In the second scenario, the offline handwriting image serves as a style image. Offline handwriting images may contain handwriting background and may contain words with variable font thickness, as shown in the leftmost column of Fig. \ref{fig:words_offline}. We evaluate our proposed method for online handwriting as listed in Table \ref{tab:overall}. HiGAN+ \cite{higanplus}, which generates state-of-the-art results for handwriting image generation, seems to have failed to imitate online sample images. It does not give satisfactory results. The poor performance of HiGAN+ on online images might be because it over-fitted during training to predict texture as well, even though there is no texture in the online style images. We also evaluated wordstylist \cite{wordstylist} and compared it with our method for online handwritten text. Although the FID score is good for the word style, the rest of the metrics are not satisfactory. The recent work on image generation vatr++ \cite{vatrpp} and one-shot diffusion model \cite{oneshot} shows comparative results for all the evaluation metrics. However, they are trained to generate images and are unable to generate stroke information. 
The stroke diffusion model \cite{vdiff} produces good results for, except it cannot replicate style well as shown in Table \ref{tab:overall}. The stroke diffusion does not follow the style template because its style features are trained on natural images \cite{mobilenet}. Our method offers an advantage over the previous techniques of handwriting style imitation since we are about to generate stroke information for the given style image and textual content. The proposed method \textit{layout + MS} gives the lower \textit{IS} and \textit{FID} scores with higher values of \textit{PNSR} and \textit{MSSIM}.
%On the other hand, both Trace \cite{trace} and stroke diffusion \cite{vdiff} are stroke sequence prediction networks. There is no issue with background texture in generated handwriting for them. They both produce satisfactory results with Trace even better in terms of \textit{PSNR} and \textit{MSSIM}. The style distance is reasonable for Trace \cite{trace} but it produces a larger projected character matching error. The lack of character shape matching is because the Trace model is based on LSTM \cite{lstm} network and trained only with \textit{DTW} loss. The \textit{DTW} loss is helpful for overall style matching but does not give sufficient importance to local character shapes. 
The Visual results are shown in Fig. \ref{fig:bdiagram}. In the given examples, we provide style image on the top of each example with textual content and Layout. We compare the effect of including \textit{layout} with and without \textit{MS} features. As we can see, the \textit{Layout + MS} gives the best results, whereas the SDM \cite{vdiff} hardly mimics the style image. We can also validate from Fig. \ref{fig:bdiagram} that the inclusion of layout information in the stroke diffusion model not only helps with proper spacing for words (via strokes) but also aids in ensuring the same number of words as in textual content by avoiding to miss words in the generated images (via strokes generation). 

%\cite{vdiff} extract style features from \cite{mobilenet} trained on natural images. The Trace \cite{trace} also extracts style features from resnet layer before feeding it into LSTM architecture for stroke prediction.

\subsection{Discussion} \label{discuss}
%Our multi-scale style feature extraction can imitate style from template images. The addition of DTW loss with \textit{MSCAP} gives us the best (lowest) projected character shape matching error as shown in the last column of Table \ref{tab:overall_offline} (projected shape).
To depict the generalization ability of our methods as compared to previous methods, we compute the evaluation metrics on offline handwriting images. These images differ from online images in terms of background texture, writing styles, and font thickness. The offline and online IAM datasets \cite{IAM-online} are composed of words and lines of text, respectively, which is another prominent difference between them. Since previous methods \cite{higanplus,wordstylist,oneshot} are trained on images of words from IAM-online datasets, we also evaluated our method against them using the same offline word images. Our stroke generation method can be applied to words as well as the long sentences of textual content. Moreover, the images from the IAM-offline dataset are completely unseen for our proposed diffusion model. 

\begin{figure*}[!ht]
    \centering
    \includegraphics[scale=0.32]{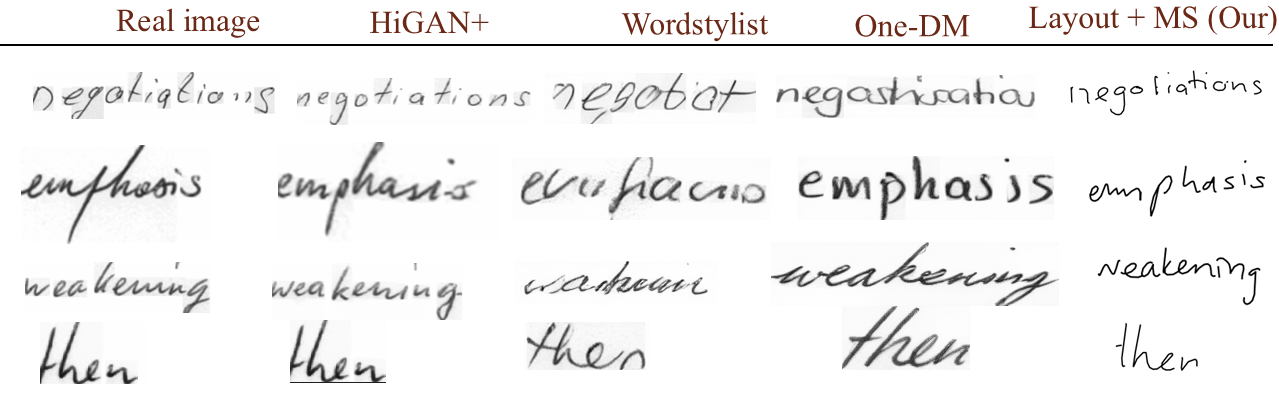}
     \caption{Visual comparison of the proposed method with the state-of-the-art handwriting image generation methods for offline handwriting word samples.} 
     \label{fig:words_offline}
\end{figure*}

For the qualitative examples shown in Fig. \ref{fig:words_offline}, the state-of-the-art HiGAN+ \cite{higanplus} produces nearly perfect results in the case of offline sample images, but HiGAN+ has poor generalizability since it does not perform reasonably on online sample images. On the other hand, \cite{wordstylist} does not extract style features from the images. Rather, it learns the style from integer input for writer ID. Therefore, \cite{wordstylist} shows the least generalization and style similarity.
\cite{oneshot} also produces good results for style imitation in most scenarios. However, our method not only shows reasonable style similarity as well as generalizability but also generates the strokes by attempting to mimic the calligraphic style from the style image. Notably, \cite{wordstylist} cannot process multiple words without modifying input interfacing. However, the proposed diffusion-based handwriting image generation method can generate short and long text without additional effort. We can also validate in Fig \ref{fig:words_offline} that the stroke generated with the diffusion model produces readable text even though we have not leveraged any text recognition module. Our proposed method is trained only on online images \cite{IAM-online}, but it can still produce competitive results for offline image samples. It shows that our diffusion model has better generalization ability than GAN architecture \cite{higanplus}. 

%\red{The description of several key components is not sufficiently clear. For example, the loss function for writer ID classification and the method for using multi-scale patch attention style features to compute the loss (Sec. 3.1.3) are not well-explained.}

%\subsubsection{Ablation study:}
%\red{It is suggested to add a separate ablation study section, you can integrate both qualitative and quantitative analyses to demonstrate the effectiveness of each module in the proposed method. Table 2 in GANwriting[j] serves as an example for reference. 
%We also compared visual results for variants. [j] GANwriting: content-conditioned generation of styled handwritten word images, ECCV, 2020. }
%\red{ The paper lacks an ablation study to validate the efficacy of the multi-scale image input and the proposed spatial and scale embeddings. }
%Our proposed method for multi-scale character pair features with DTW loss $MSCAP_{dtw}$, improves the results of FID scores, style distance, and Projected character shape matching. The multi-scale attention-based features for character pairs $MSCAP$ produce reasonable style. The DTW loss further improves the projected character shape matching error which strengthens our claim that DTW loss along with diffusion loss could improve overall handwriting stroke generation. \textit{MS}, \textit{$MS_{dtw}$}, \textit{MSCAP}, and \textit{$MSCAP_{dtw}$} of our method in Fig. \ref{fig:unconstrained}. We observe that the character shapes are good for all variants, but the overall style is better imitated for \textit{MSCAP} as we focus on character pair features.  

\begin{table*}[] 
\centering
\begin{tabular}{c|c|c|c|c}
\hline
Method            & IS $\uparrow$   & FID $\downarrow$ & PSNR  $\uparrow$ & MSSIM $\uparrow$ \\ \hline
Without layout      & 1.420 & 2.130   & 12.440 & 0.741   \\ \hline  
Layout (Ours)  & 1.517 &    1.339 & 12.611& 0.753  \\ \hline 
Layout + MS (Ours)   & 1.561 &  1.383 & 12.459 & 0.746    \\ \hline

\end{tabular}
\caption{The ablation study for our proposed method without layout and multiscale features, with layout only and layout with multi-scale style features.}
\label{tab:ablation}
\end{table*}
%It can be seen from Table \ref{tab:ablation} that the HiGAN produces good results for all the evaluation metrics except for the projected character shape matching. The reason for the extraordinary results of HiGAN+ \cite{higanplus} on the offline datasets and its inability to generate any reasonable images in the online dataset (Table \ref{tab:overall_online}) suggests that the network lacks generalization ability. It may work only on offline words, which restricts its capability to diverse style images. 
%\cite{vdiff} produces reasonable results; however, it cannot imitate style since its features are trained on natural images. 

Table \ref{tab:ablation} shows the ablation study of guiding the diffusion model with \textit{Layout}. It gives the lowest \textit{FID} scores. However, including multi-scale \textit{MS} features with \textit{Layout} (Layout +  MS) gives better \textit{PSNR} and \textit{MSSIM} matrices.
%The qualitative examples in Fig. \ref{fig:lstm_results} show that the stroke generation with proposed methods produces better stroke generation as compared to LSTM architecture \cite{trace} (Trace), stroke diffusion model \cite{vdiff}, and HiGAN+ \cite{higanplus}. \cite{vdiff} extract style features from \cite{mobilenet} trained on natural images. The Trace \cite{trace} also extracts style features from resnet layer before feeding it into LSTM architecture for stroke prediction. Whereas \cite{wordstylist} does not extract style features from the images. Rather, it learns the style from integer input for writer ID. Therefore, \cite{wordstylist} shows the least generalization and style similarity and our method \textit{$MSCAP_{dtw}$} shows the highest style similarity.
%\subsubsection{Visual comparison:}
%Some qualitative results of our method \textit{$MSCAP_{dtw}$} in comparison with the previous methods \cite{higanplus,vdiff} are shown in Fig. \ref{fig:different_text}. Here, we stress the capability of our method to perform well on unseen text content. We can see that our method ($MSCAP_{dtw}$) can imitate the unseen text well as compared to stroke diffusion \cite{vdiff}. Moreover, our method also performs well compared to the state-of-the-art image generation method (HiGAN+ \cite{higanplus}).
%\subsubsection{Time complexity:}   
Finally, we highlight the training time and auxiliary networks used in previous methods \cite{higanplus,vdiff} and our proposed method in Table \ref{tab:arch}. HiGAN+ \cite{higanplus} utilizes text recognition, patch refinement, and writer ID module. It takes 3 days on a single NVIDIA A100 GPU. The diffusion model for handwriting image generation \cite{vdiff}, does not include a text recognition module but it still takes a longer time to train due to the iterative learning of diffusion networks.
Our proposed method only takes 6 hours to complete 60k iterations to converge the learning of stroke generation with the diffusion model. Our model takes significantly less time since we generate strokes rather than images; the number of predicted strokes is much smaller than the number of pixels in the image.

\begin{table*}[]
\small
\centering
\begin{tabular}{l|c|c|c|c|c|c|c}
\hline
Method  & Params & Train time & WriterID    & Refine & Recog.  & Unseen style & Strokes \\ \hline

\multicolumn{1}{l|}{Layout + MS (Ours)}    & 16.8 M           & 24 hours      &              &                  &              & $\checkmark$   & $\checkmark$       \\ \hline
\multicolumn{1}{l|}{Wordstylist} & 40.4 M           & 7 days        &           $\checkmark$   &                  &              &                &                    \\ \hline
\multicolumn{1}{l|}{HiGAN+}      & 14.0 M           & 3 days        & $\checkmark$ & $\checkmark$     & $\checkmark$ &   &                    \\ \hline
\end{tabular}
\caption{Comparison of model size and architecture with the state-of-the-art methods.}
\label{tab:arch}
\end{table*}
%-------------------------------------------------------------------------

%-------------------------------------------------------------------------
\section{Conclusion}
We have demonstrated that the diffusion model conditioned on multi-scale improves the calligraphic style imitation for handwriting stroke generation. Importantly, we propose to include word layout, which outperforms stroke generation from sample images and produces competitive results. In our work, we explore the diffusion model guided with multi-scale features and word layout. Our method effectively generates strokes for unseen textual content and is able to imitate the handwriting style as well. Our quantitative and qualitative analysis suggests that our diffusion model can imitate various unseen handwriting styles.

%-------------------------------------------------------------------------

%
% ---- Bibliography ----
%
% BibTeX users should specify bibliography style 'splncs04'.
% References will then be sorted and formatted in the correct style.
%
% \bibliographystyle{splncs04}
% \bibliography{mybibliography}
%
\bibliographystyle{splncs04}
\bibliography{samplepaper}

\begin{thebibliography}{10}
\providecommand{\url}[1]{\texttt{#1}}
\providecommand{\urlprefix}{URL }
\providecommand{\doi}[1]{https://doi.org/#1}

\bibitem{deepwriting}
Aksan, E., Pece, F., Hilliges, O.: Deepwriting: Making digital ink editable via deep generative modeling. In: Proceedings of the 2018 CHI conference on human factors in computing systems. pp. 1--14 (2018)

\bibitem{trace}
Archibald, T., Poggemann, M., Chan, A., Martinez, T.: Trace: A differentiable approach to line-level stroke recovery for offline handwritten text. arXiv preprint arXiv:2105.11559  (2021)

\bibitem{HWT}
Bhunia, A.K., Khan, S., Cholakkal, H., Anwer, R.M., Khan, F.S., Shah, M.: Handwriting transformers. In: Proceedings of the IEEE/CVF international conference on computer vision. pp. 1086--1094 (2021)

\bibitem{bhunia2018}
Bhunia, A.K., Bhowmick, A., Bhunia, A.K., Konwer, A., Banerjee, P., Roy, P.P., Pal, U.: Handwriting trajectory recovery using end-to-end deep encoder-decoder network. In: 2018 24th International Conference on Pattern Recognition (ICPR). pp. 3639--3644. IEEE (2018)

\bibitem{diff3}
Cheng, S.I., Chen, Y.J., Chiu, W.C., Tseng, H.Y., Lee, H.Y.: Adaptively-realistic image generation from stroke and sketch with diffusion model. In: Proceedings of the IEEE/CVF Winter Conference on Applications of Computer Vision. pp. 4054--4062 (2023)

\bibitem{oneshot}
Dai, G., Zhang, Y., Ke, Q., Guo, Q., Huang, S.: One-dm: One-shot diffusion mimicker for handwritten text generation. In: European Conference on Computer Vision. pp. 410--427. Springer (2025)

\bibitem{disentangling}
Dai, G., Zhang, Y., Wang, Q., Du, Q., Yu, Z., Liu, Z., Huang, S.: Disentangling writer and character styles for handwriting generation. In: Proceedings of the IEEE/CVF conference on computer vision and pattern recognition. pp. 5977--5986 (2023)

\bibitem{hgen1}
Davis, B., Tensmeyer, C., Price, B., Wigington, C., Morse, B., Jain, R.: Text and style conditioned gan for generation of offline handwriting lines. arXiv preprint arXiv:2009.00678  (2020)

\bibitem{hgen2}
Davis, B., Tensmeyer, C., Price, B., Wigington, C., Morse, B., Jain, R.: Text and style conditioned gan for generation of offline handwriting lines. arXiv preprint arXiv:2009.00678  (2020)

\bibitem{diff2}
Dhariwal, P., Nichol, A.: Diffusion models beat gans on image synthesis. Advances in neural information processing systems  \textbf{34},  8780--8794 (2021)

\bibitem{wor}
Diaz, M., Crispo, G., Parziale, A., Marcelli, A., Ferrer, M.A.: Writing order recovery in complex and long static handwriting  (2022)

\bibitem{recog1}
Faundez-Zanuy, M., Fierrez, J., Ferrer, M.A., Diaz, M., Tolosana, R., Plamondon, R.: Handwriting biometrics: Applications and future trends in e-security and e-health. Cognitive Computation  \textbf{12}(5),  940--953 (2020)

\bibitem{higan}
Gan, J., Wang, W.: Higan: Handwriting imitation conditioned on arbitrary-length texts and disentangled styles. In: Proceedings of the AAAI Conference on Artificial Intelligence. vol.~35, pp. 7484--7492 (2021)

\bibitem{higanplus}
Gan, J., Wang, W., Leng, J., Gao, X.: Higan+: Handwriting imitation gan with disentangled representations. ACM Transactions on Graphics (TOG)  \textbf{42}(1),  1--17 (2022)

\bibitem{svit}
Han, K., Wang, Y., Chen, H., Chen, X., Guo, J., Liu, Z., Tang, Y., Xiao, A., Xu, C., Xu, Y., et~al.: A survey on vision transformer. IEEE transactions on pattern analysis and machine intelligence  \textbf{45}(1),  87--110 (2022)

\bibitem{hanif2024comprehensive}
Hanif, S.: A Comprehensive Framework for Stroke Trajectory Recovery for Unconstrained Handwritten Documents. Temple University (2024)

\bibitem{hanif_detection}
Hanif, S., Latecki, L.J.: Autonomous character region score fusion for word detection in camera-captured handwriting documents

\bibitem{u-trace}
Hanif, S., Latecki, L.J.: Strokes trajectory recovery for unconstrained handwritten documents with automatic evaluation  (2023)

\bibitem{isearch}
Hanif, S., Li, C., Alazzawe, A., Latecki, L.J.: Image retrieval with similar object detection and local similarity to detected objects. In: PRICAI 2019: Trends in Artificial Intelligence: 16th Pacific Rim International Conference on Artificial Intelligence, Cuvu, Yanuca Island, Fiji, August 26-30, 2019, Proceedings, Part III 16. pp. 42--55. Springer (2019)

\bibitem{resnet}
He, K., Zhang, X., Ren, S., Sun, J.: Deep residual learning for image recognition. In: Proceedings of the IEEE conference on computer vision and pattern recognition. pp. 770--778 (2016)

\bibitem{cascade}
Ho, J., Saharia, C., Chan, W., Fleet, D.J., Norouzi, M., Salimans, T.: Cascaded diffusion models for high fidelity image generation. The Journal of Machine Learning Research  \textbf{23}(1),  2249--2281 (2022)

\bibitem{mobilenet}
Howard, A.G., Zhu, M., Chen, B., Kalenichenko, D., Wang, W., Weyand, T., Andreetto, M., Adam, H.: Mobilenets: Efficient convolutional neural networks for mobile vision applications. arXiv preprint arXiv:1704.04861  (2017)

\bibitem{hgen3}
Kang, L., Riba, P., Rusinol, M., Fornes, A., Villegas, M.: Content and style aware generation of text-line images for handwriting recognition. IEEE Transactions on Pattern Analysis and Machine Intelligence  \textbf{44}(12),  8846--8860 (2021)

\bibitem{ganwriting}
Kang, L., Riba, P., Wang, Y., Rusinol, M., Forn{\'e}s, A., Villegas, M.: Ganwriting: content-conditioned generation of styled handwritten word images. In: Computer Vision--ECCV 2020: 16th European Conference, Glasgow, UK, August 23--28, 2020, Proceedings, Part XXIII 16. pp. 273--289. Springer (2020)

\bibitem{musiq}
Ke, J., Wang, Q., Wang, Y., Milanfar, P., Yang, F.: Musiq: Multi-scale image quality transformer. In: Proceedings of the IEEE/CVF International Conference on Computer Vision. pp. 5148--5157 (2021)

\bibitem{brush}
Kotani, A., Tellex, S., Tompkin, J.: Generating handwriting via decoupled style descriptors. In: European Conference on Computer Vision. pp. 764--780. Springer (2020)

\bibitem{layout_text_to_image}
Lakhanpal, S., Chopra, S., Jain, V., Chadha, A., Luo, M.: Refining text-to-image generation: Towards accurate training-free glyph-enhanced image generation. arXiv preprint arXiv:2403.16422  (2024)

\bibitem{vdiff}
Luhman, T., Luhman, E.: Diffusion models for handwriting generation. arXiv preprint arXiv:2011.06704  (2020)

\bibitem{IAM-online}
Marti, U.V., Bunke, H.: The iam-database: an english sentence database for offline handwriting recognition. International Journal on Document Analysis and Recognition  \textbf{5}(1),  39--46 (2002)

\bibitem{inksight}
Mitrevski, B., Rak, A., Schnitzler, J., Li, C., Maksai, A., Berent, J., Musat, C.: Inksight: Offline-to-online handwriting conversion by learning to read and write. arXiv preprint arXiv:2402.05804  (2024)

\bibitem{glide}
Nichol, A., Dhariwal, P., Ramesh, A., Shyam, P., Mishkin, P., McGrew, B., Sutskever, I., Chen, M.: Glide: Towards photorealistic image generation and editing with text-guided diffusion models. arXiv preprint arXiv:2112.10741  (2021)

\bibitem{wordstylist}
Nikolaidou, K., Retsinas, G., Christlein, V., Seuret, M., Sfikas, G., Smith, E.B., Mokayed, H., Liwicki, M.: Wordstylist: Styled verbatim handwritten text generation with latent diffusion models. arXiv preprint arXiv:2303.16576  (2023)

\bibitem{diffpen}
Nikolaidou, K., Retsinas, G., Sfikas, G., Liwicki, M.: Diffusionpen: Towards controlling the style of handwritten text generation. In: European Conference on Computer Vision. pp. 417--434. Springer (2024)

\bibitem{pred}
Nishide, S., Okuno, H.G., Ogata, T., Tani, J.: Handwriting prediction based character recognition using recurrent neural network. In: 2011 IEEE International Conference on Systems, Man, and Cybernetics. pp. 2549--2554. IEEE (2011)

\bibitem{vatr}
Pippi, V., Cascianelli, S., Cucchiara, R.: {Handwritten Text Generation from Visual Archetypes}. In: Proceedings of the IEEE/CVF Conference on Computer Vision and Pattern Recognition (2023)

\bibitem{multiseq}
Rabhi, B., Elbaati, A., Boubaker, H., Hamdi, Y., Hussain, A., Alimi, A.M.: Multi-lingual character handwriting framework based on an integrated deep learning based sequence-to-sequence attention model. Memetic Computing  \textbf{13}(4),  459--475 (2021)

\bibitem{multiauto}
Rabhi, B., Elbaati, A., Boubaker, H., Pal, U., Alimi, A.: Multi-lingual handwriting recovery framework based on convolutional denoising autoencoder with attention model  (2022)

\bibitem{dalle}
Ramesh, A., Dhariwal, P., Nichol, A., Chu, C., Chen, M.: Hierarchical text-conditional image generation with clip latents. arXiv preprint arXiv:2204.06125  \textbf{1}(2), ~3 (2022)

\bibitem{diff1}
Ramesh, A., Dhariwal, P., Nichol, A., Chu, C., Chen, M.: Hierarchical text-conditional image generation with clip latents. arXiv preprint arXiv:2204.06125  \textbf{1}(2), ~3 (2022)

\bibitem{bioinspired}
Senatore, R., Santoro, A., Parziale, A., Marcelli, A.: A biologically inspired approach for recovering the trajectory of off-line handwriting  (2022)

\bibitem{sohl}
Sohl-Dickstein, J., Weiss, E., Maheswaranathan, N., Ganguli, S.: Deep unsupervised learning using nonequilibrium thermodynamics. In: International conference on machine learning. pp. 2256--2265. PMLR (2015)

\bibitem{writelikeyou}
Tang, S., Lian, Z.: Write like you: Synthesizing your cursive online chinese handwriting via metric-based meta learning. In: Computer Graphics Forum. vol.~40, pp. 141--151. Wiley Online Library (2021)

\bibitem{vatrpp}
Vanherle, B., Pippi, V., Cascianelli, S., Michiels, N., Van~Reeth, F., Cucchiara, R.: Vatr++: Choose your words wisely for handwritten text generation. IEEE Transactions on Pattern Analysis and Machine Intelligence  (2024)

\bibitem{affganwriting}
Wang, H., Wang, Y., Wei, H.: Affganwriting: a handwriting image generation method based on multi-feature fusion. In: International Conference on Document Analysis and Recognition. pp. 302--312. Springer (2023)

\bibitem{vit}
Yuan, L., Chen, Y., Wang, T., Yu, W., Shi, Y., Jiang, Z.H., Tay, F.E., Feng, J., Yan, S.: Tokens-to-token vit: Training vision transformers from scratch on imagenet. In: Proceedings of the IEEE/CVF international conference on computer vision. pp. 558--567 (2021)

\end{thebibliography}

\end{document}